\documentclass[letterpaper, 10 pt, conference]{ieeeconf}

\IEEEoverridecommandlockouts
\usepackage{graphicx}
\usepackage{amsmath}
\usepackage{amssymb}
\usepackage{booktabs}
\usepackage{url}

\graphicspath{{figures/export/generated/}{figures/export/edited/}{figures/export/final/}}

\newcommand{\panelcaption}[1]{\par\vspace{2pt}{\footnotesize #1\par}}

\title{\LARGE \bf DUGM-R: Uncertainty-Aware Dynamic Grid Mapping and Risk-Triggered Recovery for Learned Local Navigation}
\author{
Haoyun Feng$^{1}$, Adrian Rubio-Solis$^{1}$, Zhaodong Guo$^{2}$, and George Mylonas$^{1}$
\thanks{$^{1}$The Hamlyn Centre for Robotic Surgery, Imperial College London, London, United Kingdom.}
\thanks{$^{2}$Department of Computing, Imperial College London, London, United Kingdom.}
}

\begin{document}
\maketitle
\thispagestyle{empty}
\pagestyle{empty}

\begin{abstract}
% ============================================================================
% Merged from sections/00_abstract.tex
% ============================================================================
Learned local navigation in crowded indoor environments is sensitive to how
dynamic obstacle motion is represented, while collision-prone behaviour may
persist after nominal policy training. We present a risk-aware
reinforcement-learning framework that addresses these two issues through an
uncertainty-aware Dynamic Uncertainty Grid Map (DUGM) and a modular
post-training recovery mechanism. DUGM combines local occupancy, estimated
obstacle motion, and motion-estimation uncertainty in a robot-centric
representation. After the nominal policy is frozen, a finite-horizon Risk Value
Function (RVF) is trained from nominal rollouts and used to trigger a dedicated
recovery policy when continued nominal execution is predicted to be
collision-prone. Experiments in a held-out NVIDIA Isaac Sim
clinical-logistics benchmark show that uncertainty-aware dynamic representation
improves nominal navigation over static and deterministic alternatives, while
the recovery mechanism further mitigates residual collision-prone behaviour.
The complete framework is also deployed directly on a TurtleBot3 without policy
fine-tuning, retraining, or site-specific adaptation, retaining the performance
trend observed in simulation. These results indicate that uncertainty-aware
dynamic representation and post-training recovery provide complementary
mechanisms for improving learned local navigation.
\end{abstract}

% ============================================================================
% Merged from sections/01_introduction.tex
% ============================================================================
\section{Introduction}
\label{sec:introduction}

Mobile robots can support delivery, transport, telepresence, and other tasks in
clinical-logistics and similarly constrained indoor environments.  To execute
these tasks, a robot must navigate corridors, waiting areas, offices, and
reception spaces shared with pedestrians, furniture, and mobile equipment.
Local navigation must therefore make practical short-range decisions that
balance progress and collision avoidance despite constrained geometry and
dynamic interactions \cite{morgan_robots_2022,bernhard_boosting_2024}.

Motion-aware learned local navigation can improve its response to dynamic
obstacles, but pedestrian position and velocity are estimated through online
perception and tracking rather than known exactly.  Deterministically
projecting uncertain motion estimates can consequently produce an overly
confident representation of future occupancy.  A policy-facing observation
should instead expose both estimated motion and the uncertainty attached to
that estimate, particularly when sensing is affected by noise, occlusion,
short tracks, or association errors \cite{schreiber_dynamic_2021,
thomas_learning_2022,xie_stochastic_2023}.

A second issue remains after a nominal policy has been trained:
collision-prone states can still occur at deployment.  Reward shaping,
constrained learning, and model-based safety structures can improve nominal
behaviour, but they couple safety handling to task optimisation or rely on
assumptions about dynamics and safe sets \cite{lutjens_safe_2019,
fisac_general_2019,cheng_end--end_2019}.  A complementary question is whether
an already-trained nominal policy can remain frozen and be augmented post hoc
by a risk monitor and a dedicated recovery controller
\cite{thananjeyan_recovery_2021,he_agile_2024}.

We present DUGM-R and provide empirical evidence that uncertainty-aware dynamic representation and post-training recovery offer complementary benefits for learned navigation among moving obstacles. The Dynamic Uncertainty Grid Map (DUGM) encodes local occupancy, estimated obstacle motion, and motion-estimation uncertainty in a common grid observation for the nominal policy, risk estimator, and recovery policy. After nominal training, the policy is frozen; a finite-horizon Risk Value Function (RVF) learned from its rollouts guides recovery training and triggers intervention. Experiments show improved nominal navigation over static, deterministic, and fixed-inflation representations, with further collision reduction from recovery.

The main contributions are:

\begin{itemize}
    \item We propose DUGM, which jointly represents local occupancy, estimated obstacle motion, and motion-estimation uncertainty for learned local navigation.

    \item We introduce a post-training recovery module for a frozen nominal policy. A finite-horizon RVF shapes recovery-policy training and triggers intervention using a threshold selected by validation warning performance.

    \item We evaluate DUGM-R in a held-out Isaac Sim clinical-logistics environment and demonstrate zero-shot deployment on a TurtleBot3 across unseen corridors, offices, and reception areas.
\end{itemize}

\begin{figure*}[!t]
\centering
\includegraphics[width=\textwidth]{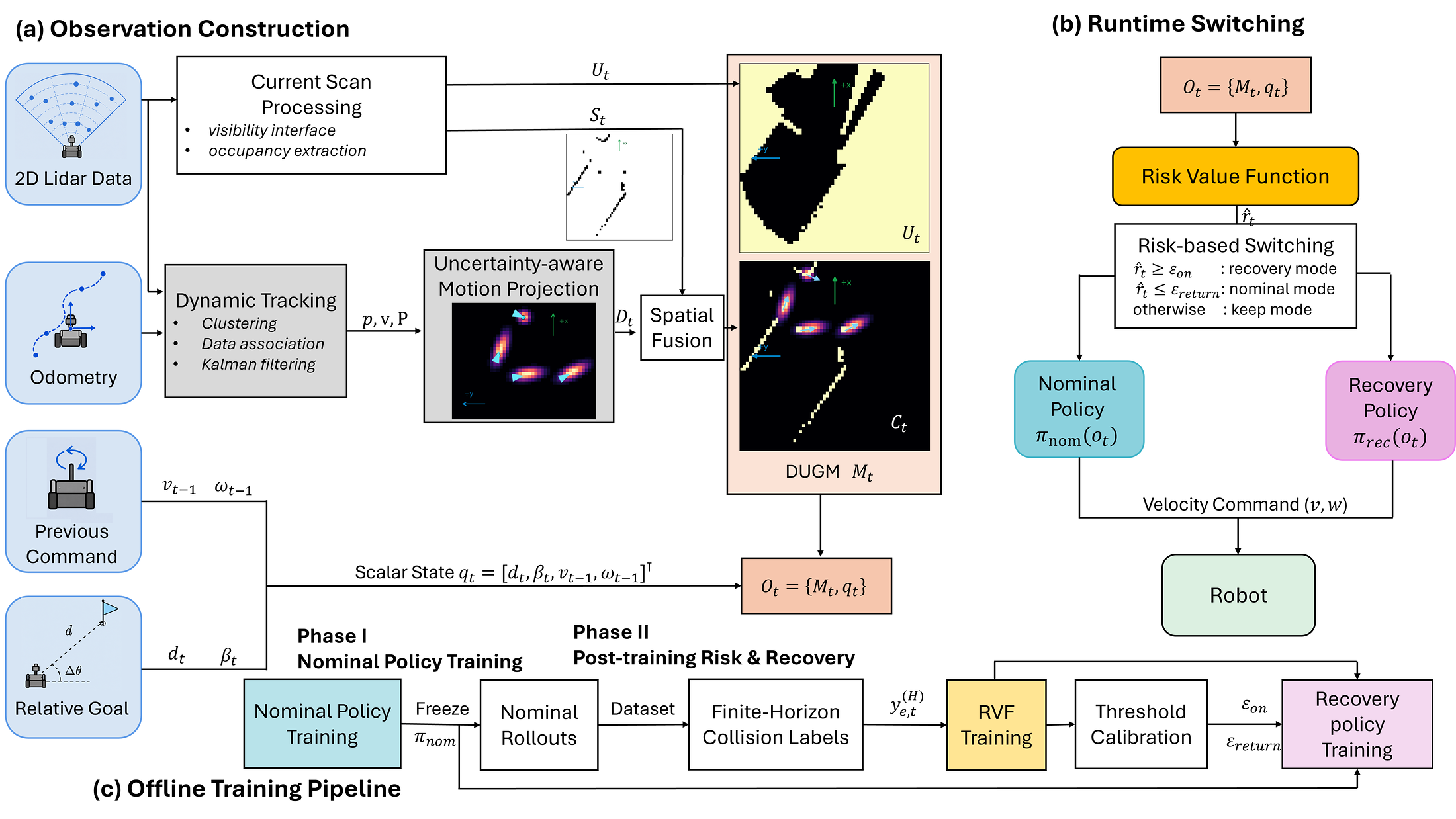}
\caption{Overview of the proposed framework:
(a) DUGM observation construction,
(b) RVF-triggered runtime switching, and
(c) the two-stage training pipeline.}
\label{fig:methods-framework-overview}
\end{figure*}

% ============================================================================
% Merged from sections/02_related_work.tex
% ============================================================================
\section{Related Work}
\label{sec:literature-review}

\subsection{Learning-Based Local Navigation and Representations}
\label{sec:related-learning-representations}

Classical local planners explicitly encode collision avoidance, kinematic
feasibility, and short-horizon motion objectives.  The Dynamic Window Approach
(DWA) searches dynamically admissible velocities \cite{fox_dynamic_1997},
whereas the Velocity Obstacle (VO) and Optimal Reciprocal Collision Avoidance
(ORCA) methods exclude velocities predicted to cause collisions
\cite{fiorini_motion_1998,van_den_berg_reciprocal_2008,
van_den_berg_reciprocal_2011}.  Trajectory-optimization methods such as the
Timed Elastic Band optimize timed local trajectories \cite{rosmann_efficient_2013},
and sampling-based Model Predictive Path Integral (MPPI) controllers evaluate
short-horizon rollouts \cite{williams_mppi_2017}.  These methods remain strong engineering
baselines, but their behaviour depends on the available obstacle states,
motion-model assumptions, planning horizon, and manually designed objectives.

To acquire more complex navigation behaviours from data or repeated
interaction, learned local navigation has explored imitation learning (IL) and
reinforcement learning (RL), particularly deep reinforcement learning (DRL).
Map-based IL learns obstacle
avoidance from expert demonstrations \cite{liu_map-based_2018}.  DRL methods
have learned socially aware motion policies \cite{chen_socially_2017},
navigation among dynamic decision-making agents \cite{everett_motion_2018},
attention-based crowd interaction \cite{chen_crowd-robot_2019}, and map-based
local navigation \cite{chen_robot_2020,yao_crowd-aware_2021}.  Other work
combines learned policies with classical local-planning structure, as in
DWA-RL \cite{patel_dwa-rl_2021}.  Once the local controller is learned, the
observation representation becomes a central modelling choice.

At the sensor level, learned policies can consume light detection and ranging
(LiDAR) scans or short temporal
range histories directly \cite{xie_towards_2021,xie_drl-vo_2023}.  Such inputs
retain measurement detail and require little object-specific abstraction, but
place the burden of extracting relevant geometry and motion on the learned
encoder.  Agent-level representations instead expose nearby pedestrians
through quantities such as relative position, velocity, heading, radius, and
interaction features \cite{chen_socially_2017,everett_motion_2018}.  Attention
mechanisms can aggregate a variable number of such agent states
\cite{chen_crowd-robot_2019}, although the resulting policy depends on reliable
detection, association, tracking, and state estimation.  DRL-VO combines a
short LiDAR history and pedestrian kinematics with a VO-shaped learning
objective \cite{xie_drl-vo_2023}, illustrating how explicit dynamic-interaction
structure can inform a learned local controller while retaining this dependence
on the perception and tracking pipeline.

Grid-based representations provide a spatial interface between perception and
policy.  Occupancy grids encode robot-centric free and occupied space while
abstracting object identity \cite{elfes_using_1989}; related learned-navigation
methods use local maps or pedestrian maps for IL and DRL
\cite{liu_map-based_2018,chen_robot_2020,yao_crowd-aware_2021}.  Dynamic
occupancy grids extend this interface with motion estimates
\cite{schreiber_dynamic_2021}, while learned spatiotemporal maps predict future
occupancy \cite{thomas_learning_2022}.  Stochastic occupancy prediction further
represents multiple possible scene futures \cite{xie_stochastic_2023}.  These
works show that explicit dynamic and predictive representations can expose
valuable interaction information; once future occupancy is predicted, however,
uncertainty in sensing, tracking, motion estimation, and future evolution
becomes increasingly consequential.

\subsection{Safety and Recovery for Learned Navigation}
\label{sec:related-safety-recovery}

Safety can be incorporated during nominal-policy learning through
collision-aware reward shaping \cite{xie_towards_2021,xie_drl-vo_2023} or by
optimizing task reward subject to explicit cost constraints, as in Constrained
Policy Optimization \cite{achiam_constrained_2017}.  These strategies can
improve nominal behaviour, but safety and task performance remain coupled in
the training problem and changes to their trade-off generally require further
policy optimization.

Model-based safety mechanisms instead intervene around a learned controller.
Hamilton--Jacobi reachability can support runtime safety filtering under
uncertain dynamics \cite{fisac_general_2019}, while control barrier functions
can enforce state constraints during policy learning and execution
\cite{cheng_end--end_2019}.  Their guarantees and applicability depend on the
assumed dynamics, disturbance bounds, safe sets, or tractability of the
resulting constraints.

Recovery RL pretrains a safety critic and recovery policy, then updates them alongside the task policy during interaction \cite{thananjeyan_recovery_2021}. ABS uses reach-avoid values to optimize commands tracked by a recovery policy; its static-obstacle value training limits the reported dynamic generalization to quasi-static scenes \cite{he_agile_2024}. Motivated by this limitation, DUGM explicitly encodes obstacle motion and estimation uncertainty for nominal navigation, risk assessment, and recovery. Our nominal policy remains frozen while a finite-horizon RVF learned from its rollouts directly shapes recovery training.

% ============================================================================
% Merged from sections/03_method.tex
% ============================================================================

\section{Method}
\label{sec:methods}

The proposed controller consists of a nominal policy and a post-training
recovery layer.  The nominal policy operates on a robot-centric DUGM.  After
training, the nominal policy is frozen and rolled out to train a finite-horizon
RVF.
The RVF estimates the risk of continued nominal execution and activates a
separate recovery policy when that risk is high.  The overall pipeline is
illustrated in Fig.~\ref{fig:methods-framework-overview}.

\subsection{DUGM Observation Representation}
\label{sec:methods-observation}

As shown in Fig.~1(a), DUGM is constructed from the current LiDAR scan
through three successive stages: dynamic tracking, uncertainty-aware motion
prediction, and spatial fusion. The resulting robot-centric grid is combined
with a four-dimensional scalar state $q_t$, containing normalized goal
distance and bearing together with the previously executed linear and angular
velocities, to form the policy observation.

The current-scan branch converts the LiDAR observation into a
$16\,\mathrm{m}\times16\,\mathrm{m}$ robot-centric grid at
$0.25\,\mathrm{m}$ resolution. It provides a current scan endpoint map
$S_t$ and an unsafe-space map $U_t$, in which occupied and currently
unobserved cells are treated as unsafe. Invalid LiDAR returns do not create
free-space evidence, so unobserved cells remain explicitly unsafe.

Dynamic tracks are obtained from LiDAR clustering, inter-frame association,
and a constant-velocity Kalman filter (KF); simulator ground-truth pedestrian
states are never provided to the policy. Track states are transformed into
the current robot frame before prediction. This stage corresponds to the
Dynamic Tracking block in Fig.~1(a) and provides the estimated obstacle states
and their uncertainty for subsequent motion prediction.

For track $m$ with estimated position $p_{t,m}$, velocity $v_{t,m}$, and
speed-clipped velocity $\bar{v}_{t,m}$, we first linearly project its expected
centre over the prediction horizon:
\begin{equation}
\mu_{t,m}(\tau)=p_{t,m}+\tau\bar{v}_{t,m}.
\label{eq:dugm-predicted-centre}
\end{equation}
The reported rollout uses a $4.0\,\mathrm{s}$ horizon sampled every
$0.2\,\mathrm{s}$, with pedestrian speed capped at
$0.70\,\mathrm{m\,s^{-1}}$.

The linear rollout specifies where an obstacle is expected to move, but
treating this trajectory deterministically would ignore uncertainty in the
tracked state and future motion. DUGM therefore propagates this uncertainty
over the same prediction horizon, corresponding to the Uncertainty-aware
Motion Prediction block in Fig.~1(a). To represent this uncertainty spatially,
we express the predicted position covariance in a motion-aligned basis.

Let $B=[e_{\parallel},e_{\perp}]$ be the basis aligned with the estimated
track velocity, with an isotropic branch for speeds below
$0.05\,\mathrm{m\,s^{-1}}$. The KF position covariance is propagated to
time $\tau$ and expressed in this basis as
$\sigma^2_{\mathrm{KF},\parallel}(\tau)$ and
$\sigma^2_{\mathrm{KF},\perp}(\tau)$. The directional position variances used
by the rasterizer are
\begin{equation}
\tilde{\sigma}_{\parallel,\perp}^{2}(\tau)
=
\sigma_{\mathrm{KF},\parallel,\perp}^{2}(\tau)
+\frac{1}{4}\bigl(\sigma_a s^a_{\parallel,\perp}\bigr)^2\tau^4
+\epsilon_{\Sigma},
\label{eq:dugm-directional-variance}
\end{equation}
where the nominal acceleration prior is
$\sigma_a=0.50\,\mathrm{m\,s^{-2}}$, the anisotropic scales are
$s^a_{\parallel}=1.0$ and $s^a_{\perp}=0.15$, and
$\epsilon_{\Sigma}=10^{-4}\,\mathrm{m^2}$. Thus, the predicted uncertainty
combines online state-estimation uncertainty from the Kalman filter with
increasing uncertainty due to unmodelled future acceleration.

The effective covariance is
\begin{equation}
\begin{aligned}
\Sigma^{\mathrm{eff}}_{t,m}(\tau)
&=
B
\begin{bmatrix}
\sigma_{\parallel}^{2}(\tau) & 0\\
0 & \sigma_{\perp}^{2}(\tau)
\end{bmatrix}
B^\top,\\
\sigma_d^2(\tau)
&=
\operatorname{clip}\!\left(
\tilde{\sigma}_d^2(\tau),
\lambda_{\min},
\lambda_d^{\max}
\right),
\quad d\in\{\parallel,\perp\}.
\end{aligned}
\label{eq:dugm-effective-covariance}
\end{equation}
with $\lambda_{\min}=10^{-4}\,\mathrm{m^2}$,
$\lambda_{\parallel}^{\max}=0.64\,\mathrm{m^2}$, and
$\lambda_{\perp}^{\max}=0.09\,\mathrm{m^2}$. The implementation also scales
the velocity covariance by $s^v_{\parallel}=1.0$ and
$s^v_{\perp}=0.25$ before propagation. The covariance grows anisotropically with the prediction horizon up to the specified caps, which limit excessive spatial expansion under prolonged acceleration extrapolation. Each projected obstacle state is thus
represented by a mean--covariance pair
$(\mu_{t,m}(\tau),\Sigma^{\mathrm{eff}}_{t,m}(\tau))$, which is then
rasterized into the dynamic prediction map.

To convert these continuous predictions into the policy grid, each
mean--covariance pair defines an anisotropic Gaussian spatial support.
For cell centre $g_{ij}$, the dynamic prediction is rasterized using the
Mahalanobis distance:
\begin{equation}
\begin{aligned}
\delta^2_{t,m,\tau}(i,j)
&=
(g_{ij}-\mu_{t,m}(\tau))^\top\\[-1mm]
&\quad (\Sigma^{\mathrm{eff}}_{t,m}(\tau))^{-1}
 (g_{ij}-\mu_{t,m}(\tau)),\\
D_t(i,j)
&=
\max_{(m,\tau)\in\mathcal{S}_t}
\exp\!\left(-\tfrac{1}{2}\delta^2_{t,m,\tau}(i,j)\right).
\end{aligned}
\label{eq:dugm-gaussian-diffusion}
\end{equation}
Only track--time pairs satisfying
$\delta^2_{t,m,\tau}(i,j)\leq\chi^2_{\max}$ are retained in $\mathcal{S}_t$.
Here, $\chi^2_{\max}=6.25$, and the value is zero outside this support. Taking
the maximum over all retained tracks and prediction times produces the dynamic
prediction map $D_t$.

Finally, the Spatial Fusion stage in Fig.~1(a) combines the predicted dynamic
support $D_t$ with the current LiDAR endpoint map $S_t$:
\begin{equation}
C_t(i,j)=\max\{S_t(i,j),D_t(i,j)\}.
\label{eq:dugm-visible-cost-main}
\end{equation}
The fused result forms the second DUGM channel, while the first channel retains
the current unsafe-space representation. The final two-channel
$64\times64$ robot-centric grid is therefore
\begin{equation}
M_t[0,i,j]=U_t(i,j),
\qquad
M_t[1,i,j]=C_t(i,j).
\label{eq:obs-grid-branch}
\end{equation}
Together with the scalar state $q_t$, this grid forms the complete observation
supplied to the nominal policy.

This construction also defines the distinctions used in the observation
ablation. The occupancy grid map (OGM) retains only static occupancy, while a
velocity-augmented variant (OGM-V2) adds sparse velocity channels.
Deterministic dynamic grid mapping (DDGM) uses the same tracked states and
deterministic motion rollout as DUGM, but does not propagate estimation
uncertainty into the predicted spatial support. DDGM+FI additionally applies fixed spatial inflation matched to DUGM's clipping upper bounds in Eq.~\eqref{eq:dugm-effective-covariance}.
% Deliberately introduced here so that this single-column float follows the
% opening DUGM formulation instead of fragmenting the Method opening page.
\begin{figure}[!t]
\centering
\includegraphics[width=\columnwidth]{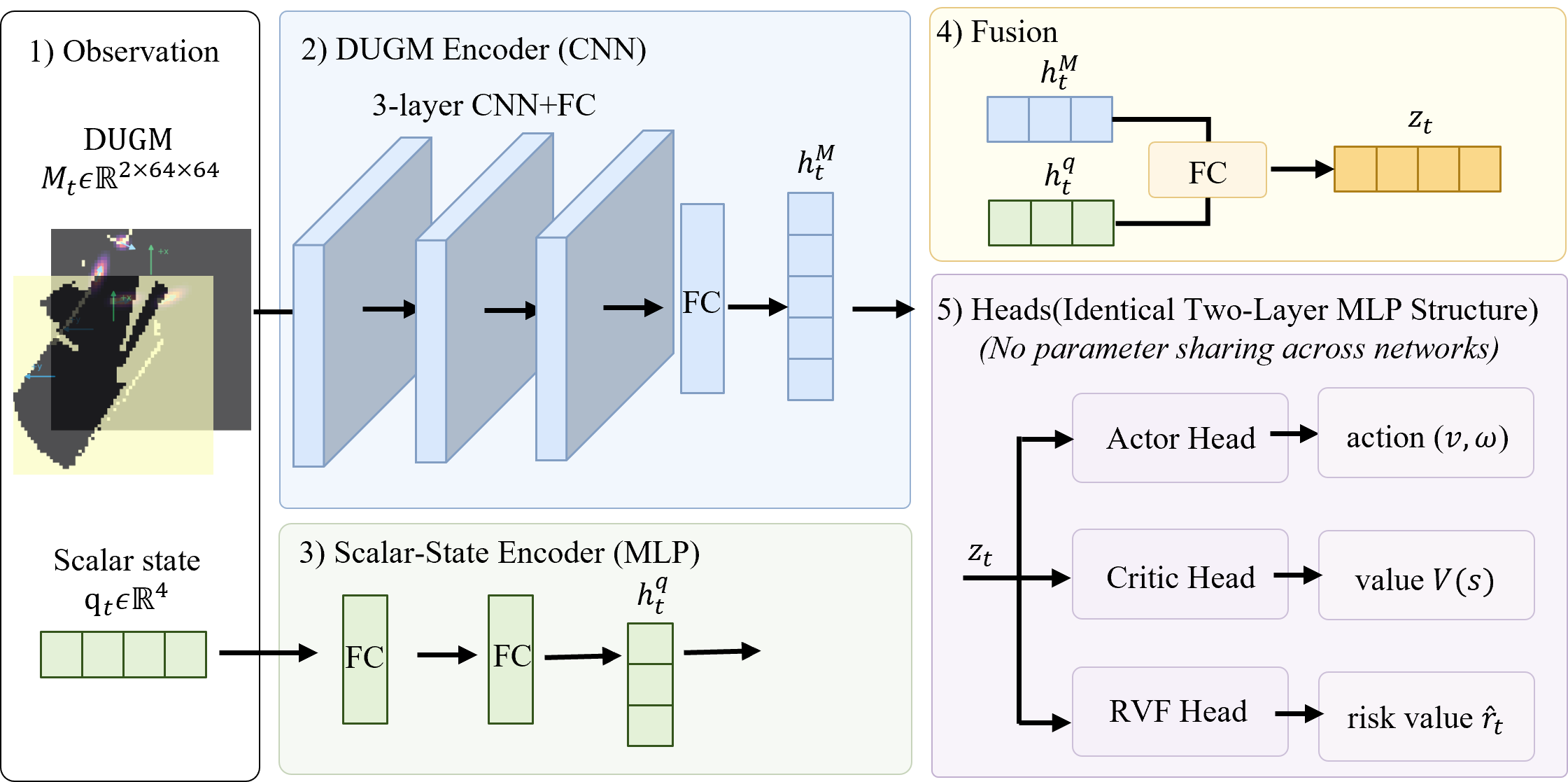}
\caption{Architecture template for the nominal policy, recovery policy, and RVF. Each network has an independent instance of blocks 1–4 and its corresponding output heads; no parameters are shared across networks.}
\label{fig:methods-network-structure}
\end{figure}

\subsection{Nominal Policy}
\label{sec:methods-nominal-policy}

The nominal policy maps the DUGM observation to a continuous
linear--angular velocity action
$a_t=(a_t^v,a_t^\omega)\in[-1,1]^2$, which is scaled to
$v_t\in[0,0.22]\,\mathrm{m\,s^{-1}}$ and
$\omega_t\in[-2.84,2.84]\,\mathrm{rad\,s^{-1}}$.
Executed commands are rate-limited to ensure smooth and feasible actuation.
The policy uses the grid--scalar processing architecture illustrated in
Fig.~\ref{fig:methods-network-structure} and is trained with proximal policy
optimization (PPO). The reward is designed to balance
three objectives: goal-directed task completion, collision-avoidance shaping,
and smooth executable motion. It is decomposed as
\begin{equation}
\begin{aligned}
r_t^{\mathrm{nom}}={}&r_t^{\mathrm{prog}}+r_t^{\mathrm{goal}}
+r_t^{\mathrm{col}}+r_t^{\mathrm{timeout}}\\
&+r_t^{\mathrm{near}}+r_t^{\mathrm{close}}+r_t^{\mathrm{reg}}.
\end{aligned}
\label{eq:nominal-total-reward}
\end{equation}

To provide dense feedback toward the local goal, the progress term is
\begin{equation}
r_t^{\mathrm{prog}}=4.0(d_t-d_{t+1}),
\label{eq:nominal-progress-reward}
\end{equation}
where $d_t$ and $d_{t+1}$ are the pre- and post-action distances to the
local goal. Episode-level success and failure are specified separately through
the terminal terms
\begin{equation}
\begin{aligned}
r_t^{\mathrm{goal}}&=60\,\mathbb{I}[d_{t+1}\le0.25\,\mathrm{m}],\\
r_t^{\mathrm{col}}&=-100\,\mathbb{I}[\mathrm{collision}_{t+1}],\\
r_t^{\mathrm{timeout}}&=-40\,\mathbb{I}[\mathrm{timeout}_{t+1}].
\end{aligned}
\label{eq:nominal-terminal-rewards}
\end{equation}

To discourage unsafe interactions before an actual collision occurs, obstacle
proximity and closing motion are shaped separately. Let $c_t$ denote the
minimum clearance after the corresponding control step. The near-obstacle
term is
\begin{equation}
\begin{aligned}
r_t^{\mathrm{near}}&=-0.125\,\phi_{\mathrm{near}}(c_{t+1}),\\
\phi_{\mathrm{near}}(c)&=\operatorname{clip}\!\left(
\frac{1.2-c}{1.2-0.101},0,1\right).
\end{aligned}
\label{eq:nominal-near-reward}
\end{equation}
and the closing-clearance term is
\begin{equation}
\begin{aligned}
r_t^{\mathrm{close}}&=-0.1\,\phi_{\mathrm{close}}(c_t,c_{t+1}),\\
\phi_{\mathrm{close}}(c_t,c_{t+1})&=\operatorname{clip}\!\left(
\frac{\max(0,c_t-c_{t+1})}{0.22\,\Delta t},0,1\right).
\end{aligned}
\label{eq:nominal-close-reward}
\end{equation}
The proximity term penalizes small clearance below $1.2\,\mathrm{m}$,
whereas the closing term penalizes a reduction in clearance between consecutive
steps, with $\Delta t=0.1\,\mathrm{s}$. The $0.101\,\mathrm{m}$ value is a
reward-shaping anchor and is not the Isaac Sim episode collision threshold.

Finally, to encourage smooth and executable commands without dominating the
primary navigation objective, we add weak regularization terms:
\begin{equation}
\begin{aligned}
r_t^{\mathrm{reg}}={}&-0.04-0.1\max(|\omega_t|-1.2,0)
-0.02\,\mathbb{I}[v_t\le0.05]\\
&-0.05\frac{|v_t-v_{t-1}|}{0.60\Delta t}
-0.05\frac{|\omega_t-\omega_{t-1}|}{7.50\Delta t}.
\end{aligned}
\label{eq:nominal-regularization}
\end{equation}

The nominal policy uses no separate dynamic-risk reward conditioned on the
action. Estimated pedestrian motion and its uncertainty enter the policy only
through DUGM.

\subsection{Finite-Horizon RVF}
\label{sec:methods-risk-value}

Although the nominal policy is trained for collision avoidance, residual
collision-prone states may still occur at deployment. Following the general
idea of separating task execution from learned risk assessment and recovery
in Recovery RL~\cite{thananjeyan_recovery_2021}, we learn a separate RVF after
freezing the nominal policy. The RVF estimates the
risk associated with continuing nominal execution over a finite horizon and
provides the trigger signal for the recovery policy described in
Sec.~\ref{sec:methods-recovery}.

After the nominal policy is frozen, 50,000 deterministic nominal-policy
episodes are collected without recovery. We use a horizon of $H=40$ control
steps, corresponding to approximately $4\,\mathrm{s}$ at the
$10\,\mathrm{Hz}$ control rate. For an episode ending at $T_e$, we define the
finite-horizon collision return as
\begin{equation}
G_t^{\mathrm{risk}}
=
\mathbb{I}\!\left[
\exists u\in[t,\min(t+H,T_e)]:
c_u^{\mathrm{col}}=1
\right],
\label{eq:risk-finite-horizon-return}
\end{equation}
and the corresponding risk value under continued nominal execution as
\begin{equation}
V_R^{\pi_{\mathrm{nom}}}(o_t)
=
\mathbb{E}_{\pi_{\mathrm{nom}}}
\left[
G_t^{\mathrm{risk}}\mid o_t
\right].
\label{eq:risk-value-function}
\end{equation}
Each nominal rollout therefore provides a Monte Carlo target for the
finite-horizon risk value. Timeout is not treated as a collision-positive
event unless a collision occurs within the prediction horizon.

The RVF takes the same grid and scalar observation as the nominal policy, but is
trained as an independent network with no shared weights or gradients, as
illustrated in Fig.~\ref{fig:methods-network-structure}. It outputs a scalar risk
score and is trained from the rollout-derived targets using class-weighted
binary cross-entropy. Because class weighting is used and no post-hoc calibration is
applied, its output $\hat r_t$ is treated as an uncalibrated finite-horizon
risk score rather than a calibrated collision probability or a formal safety
certificate.

\begin{figure}[!t]
\centering

\begin{minipage}[b]{0.49\columnwidth}
  \centering
  \includegraphics[width=\linewidth]{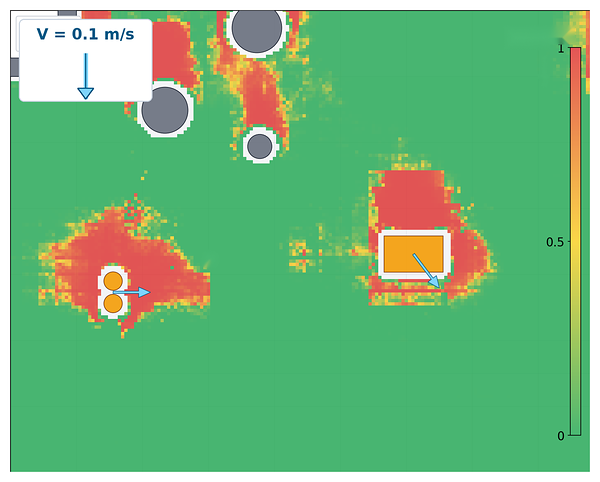}
  \panelcaption{(a) Down, $0.1\,\mathrm{m\,s^{-1}}$.}
\end{minipage}\hfill
\begin{minipage}[b]{0.49\columnwidth}
  \centering
  \includegraphics[width=\linewidth]{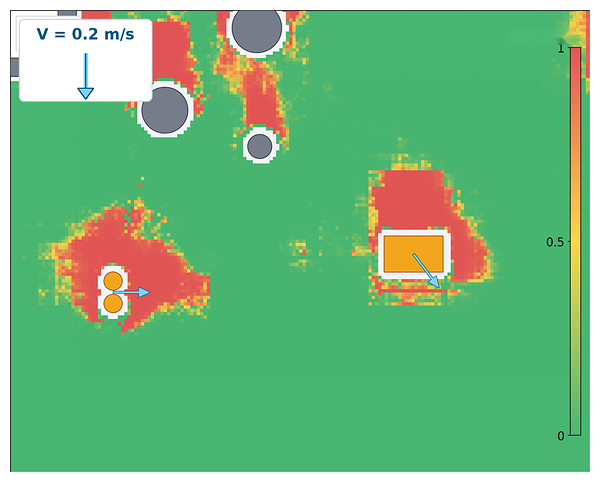}
  \panelcaption{(b) Down, $0.2\,\mathrm{m\,s^{-1}}$.}
\end{minipage}

\begin{minipage}[b]{0.49\columnwidth}
  \centering
  \includegraphics[width=\linewidth]{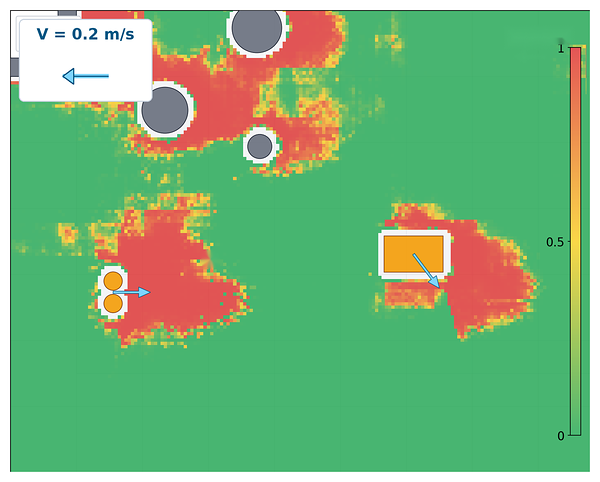}
  \panelcaption{(c) Left, $0.2\,\mathrm{m\,s^{-1}}$.}
\end{minipage}\hfill
\begin{minipage}[b]{0.49\columnwidth}
  \centering
  \includegraphics[width=\linewidth]{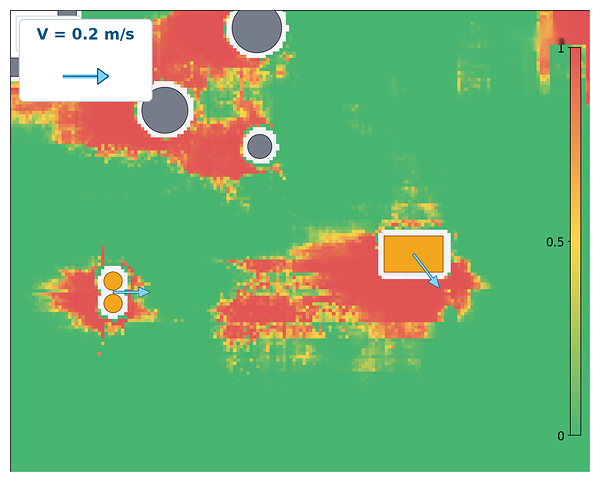}
  \panelcaption{(d) Right, $0.2\,\mathrm{m\,s^{-1}}$.}
\end{minipage}

\caption{RVF scores queried across robot positions in a fixed scene. Green/red indicate low/high scores; blue arrows show motion directions. Top-left insets specify the robot velocity used in each panel.}
\label{fig:rvf-spatial-sweep}
\end{figure}

To provide intuition for the learned risk representation,
Fig.~\ref{fig:rvf-spatial-sweep} visualizes the RVF response in a fixed scene
while varying only the robot motion state. The direction- and
speed-dependent score fields illustrate that the RVF captures the risk of
continued nominal execution rather than encoding static obstacle proximity
alone.

Before deployment, activation thresholds are selected on the held-out nominal-rollout validation split. Candidate thresholds must satisfy $\mathrm{Recall}\geq0.90$, $\mathrm{FPR}\leq0.25$, and a mean collision-warning time $W\geq0.5\,\mathrm{s}$, after which lower false-intervention rates are preferred. For DUGM, this yields $\epsilon_{\mathrm{on}}=0.634$ (90.0\% recall, 1.14\% FPR) and $\epsilon_{\mathrm{return}}=0.317=0.5\epsilon_{\mathrm{on}}$; other representations are calibrated independently using the same procedure.

\subsection{Recovery Policy and Runtime Switching}
\label{sec:methods-recovery}

The recovery policy is trained with PPO after both the nominal policy and
the RVF are frozen. It uses the same observation and action interfaces as the nominal policy, but
is trained with an independent policy network. Unlike the nominal policy, recovery is not intended to
continuously maximize goal progress; its immediate objective is to move the
robot away from collision-prone states while preserving the possibility of
eventual task completion. Accordingly, the dense goal-progress term used by
the nominal policy is replaced by an RVF-based risk-reduction signal, while
terminal task outcomes, local collision-avoidance shaping, and action
regularization are retained. The recovery reward is
\begin{equation}
\begin{aligned}
r_t^{\mathrm{rec}}={}&r_t^{\mathrm{risk}}+r_t^{\mathrm{goal}}
+r_t^{\mathrm{col}}+r_t^{\mathrm{timeout}}\\
&+r_t^{\mathrm{handoff}}+r_t^{\mathrm{clear}}
+r_t^{\mathrm{fwd}}+r_t^{\mathrm{reg}}.
\end{aligned}
\label{eq:recovery-total-reward}
\end{equation}

Because recovery is activated specifically in high-risk states, its primary
dense objective is to move toward states assigned lower risk by the frozen
RVF. Let $\hat r_t$ denote the RVF score. Risk reduction is rewarded by
\begin{equation}
r_t^{\mathrm{risk}}
=
\alpha(\hat r_t-\hat r_{t+1}),
\qquad \alpha=1.0.
\label{eq:recovery-risk-reward}
\end{equation}
A transition therefore receives positive reward when the estimated
finite-horizon risk decreases. The RVF remains fixed throughout recovery
training, so the recovery policy is optimized against a stationary risk
signal rather than co-adapting the risk estimator.

Risk reduction alone, however, does not ensure that recovery leaves the robot
in a state from which the navigation task can still be completed. The dense
goal-progress reward is therefore omitted, but final task outcomes are retained
to discourage recovery behaviours that reduce risk at the expense of eventual
task completion:
\begin{equation}
\begin{aligned}
r_t^{\mathrm{goal}}
&=8\,\mathbb{I}[\mathrm{final\ success}],\\
r_t^{\mathrm{col}}
&=-12\,\mathbb{I}[\mathrm{final\ collision}],\\
r_t^{\mathrm{timeout}}
&=-4\,\mathbb{I}[\mathrm{final\ timeout}].
\end{aligned}
\label{eq:recovery-terminal-reward}
\end{equation}
Recovery must also return control only when nominal execution can continue
successfully. A failed handoff back to the nominal policy is therefore penalized by
\begin{equation}
r_t^{\mathrm{handoff}}
=
-4\,\mathbb{I}[\mathrm{handoff\ failure}],
\label{eq:recovery-handoff-reward}
\end{equation}
where a handoff is considered unsuccessful if the subsequent nominal
execution terminates in collision or timeout.

In addition to the learned RVF signal, direct geometric shaping is retained
to preserve basic close-range collision avoidance. Let $\kappa_{t+1}$ denote
the post-action minimum clearance. Small clearance is penalized by
\begin{equation}
r_t^{\mathrm{clear}}
=
-0.25\left(
\frac{\max(0,0.35-\kappa_{t+1})}{0.35}
\right)^2,
\label{eq:recovery-clearance-reward}
\end{equation}
while continued forward motion when already close to an obstacle is
discouraged by
\begin{equation}
r_t^{\mathrm{fwd}}
=
-0.20\,
\mathbb{I}[
\kappa_{t+1}<0.35\,\mathrm{m}
\wedge v_t>0
]v_t.
\label{eq:recovery-forward-reward}
\end{equation}
These terms complement the finite-horizon RVF with an explicit local
clearance signal during corrective manoeuvres.

Finally, to encourage smooth recovery actions, weak action-magnitude and
action-change penalties are applied:
\begin{equation}
r_t^{\mathrm{reg}}
=
-0.01
-0.005\lVert a_t\rVert_2^2
-0.01\lVert a_t-a_{t-1}\rVert_2^2.
\label{eq:recovery-regularization}
\end{equation}

Only recovery-controlled transitions are stored in the recovery PPO buffer.
After control is handed back to the nominal policy, it resumes execution
until the next recovery trigger or a terminal event. If this nominal
continuation terminates the episode, the corresponding task outcome and
handoff signal are assigned to the immediately preceding recovery transition;
nominal-policy transitions themselves are not inserted into the recovery
buffer.

At deployment, the frozen RVF and the two independently trained policies are
combined through the hysteretic switching mechanism illustrated in
Fig.~1(b):
\begin{equation}
\pi_t=
\begin{cases}
\pi_{\mathrm{rec}}, & \hat r_t\ge\epsilon_{\mathrm{on}},\\
\pi_{\mathrm{nom}}, & \hat r_t\le\epsilon_{\mathrm{return}},\\
\pi_{t-1}, & \text{otherwise}.
\end{cases}
\label{eq:dugm-switching-rule}
\end{equation}
Here, $\pi_{\mathrm{nom}}$ and $\pi_{\mathrm{rec}}$ denote the nominal and
recovery policies, respectively. The lower return threshold suppresses rapid
nominal--recovery switching and
allows control to return to the nominal policy only after the RVF score has
fallen sufficiently below the activation threshold.

% ============================================================================
% Merged from sections/04_experimental_setup.tex
% ============================================================================
\section{Experiments}
\label{sec:experiments}

\subsection{Experimental Setup}
\label{sec:experiments-setup}

Policies were trained in a lightweight procedural 2-D simulator containing
straight, L-shaped, and T-junction corridors with random static obstacles and
three to five moving pedestrians. Corridor geometry, start--goal
configurations, pedestrian motion, and LiDAR noise were randomised at reset.
Training and deployment used the same TurtleBot3 footprint, 360$^\circ$ LiDAR
interface, $0.1\,\mathrm{s}$ control interval, and command limits.

\begin{figure}[!t]
\centering
\begin{minipage}[t]{0.49\columnwidth}
  \centering
  \includegraphics[width=\linewidth]{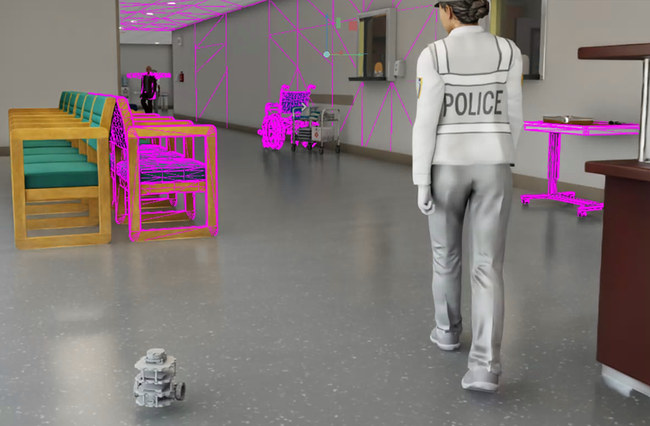}
  \panelcaption{(a) Collision and sensing model.}
\end{minipage}\hfill
\begin{minipage}[t]{0.49\columnwidth}
  \centering
  \includegraphics[width=\linewidth]{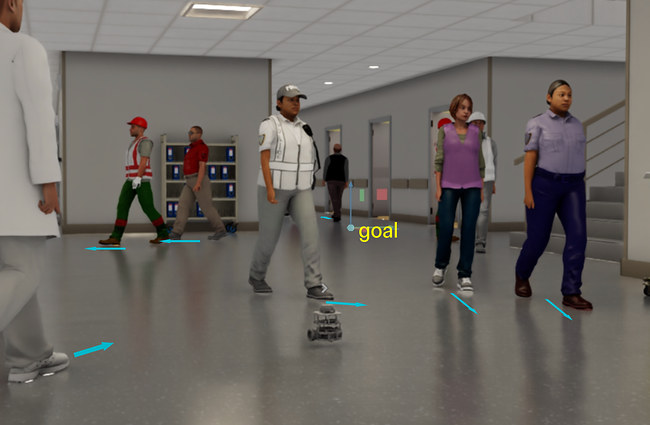}
  \panelcaption{(b) Dense pedestrian interaction.}
\end{minipage}
\caption{Held-out Isaac Sim clinical-logistics benchmark used for final
evaluation, including static furnishings and dynamic pedestrian interactions.}
\label{fig:clinical-logistics-test-environment}
\end{figure}

Final evaluation was conducted in a held-out NVIDIA Isaac Sim
clinical-logistics environment containing interconnected corridors, hospital
furnishings, and animated pedestrians. Static and dynamic objects were included
in the raycast and collision model. Three crowd-density conditions contributed
30 fixed start--goal tasks each, giving 90 episodes per method. An episode was
successful within $0.25\,\mathrm{m}$ of the goal, timed out after $200\,\mathrm{s}$,
and was marked as a collision when the minimum valid PhysX raycast hit distance
to obstacle collision geometry was at most $0.19\,\mathrm{m}$. With the
$0.089\,\mathrm{m}$ robot radius, this corresponds to approximately
$0.10\,\mathrm{m}$ of remaining surface clearance. Representative benchmark
views are shown in
Fig.~\ref{fig:clinical-logistics-test-environment}.

All methods were evaluated on the same 90 held-out tasks under identical
sensing, actuation, and terminal conditions. DWA, Nav2 MPPI, ORCA, and DRL-VO
were adapted only to the common local-navigation interface and task contract;
their implementation hyperparameters were retained, with no benchmark-specific
tuning. OGM--Lagrangian was included as an in-project constrained-RL safety
comparison. ORCA and DRL-VO received dynamic-obstacle estimates from the same
LiDAR clustering and Kalman-tracking pipeline as the proposed method rather
than simulator ground truth.

A nominal policy using LiDAR history is included in the recovery ablation. Its
observation concatenates four consecutive LiDAR frames, ordered from oldest to
newest, with the current goal and previous-command scalars.

\subsection{Evaluation Metrics}
\label{sec:experiments-metrics}

We report success rate (SR), collision rate (CR), timeout rate (TR), and the
mean executed linear velocity over successful episodes
($\bar{v}$). Outcome rates are computed over all evaluation
episodes, while $\bar{v}$ provides a supplementary measure of
motion conservatism.

% ============================================================================
% Merged from sections/05_results.tex
% ============================================================================
\subsection{Simulation Results}
\label{sec:experiments-benchmark-results}

Table~\ref{tab:simulation_main} reports the external baseline comparison
together with ablations on observation representation and post-training
recovery.

\begin{table}[t]
\centering
\setlength{\tabcolsep}{1.5pt}
\renewcommand{\arraystretch}{1.05}
\caption{Simulation results over the 90 held-out benchmark episodes.}
\label{tab:simulation_main}
\begin{tabular}{@{}llcccc@{}}
\toprule
Group & Method & SR $\uparrow$ & CR $\downarrow$ & TR $\downarrow$ &
$\bar{v}$ (m/s) $\uparrow$ \\
\midrule
External & DWA & 51.1\% & 15.6\% & 33.3\% & 0.150 \\
& Nav2 MPPI & 35.6\% & \textbf{3.3\%} & 61.1\% & 0.114 \\
& ORCA & 27.8\% & 65.6\% & 6.6\% & 0.125 \\
& DRL-VO & 38.9\% & 61.1\% & 0.0\% & 0.134 \\
& OGM--Lagrangian & 58.9\% & 41.1\% & 0.0\% & 0.140 \\
\midrule
Observation & OGM & 57.8\% & 41.1\% & 1.1\% & 0.149 \\
& OGM-V2 & 62.2\% & 37.8\% & 0.0\% & 0.132 \\
& DDGM & 52.2\% & 47.8\% & 0.0\% & 0.136 \\
& DDGM+FI & 54.4\% & 45.6\% & 0.0\% & 0.157 \\
& DUGM & \textbf{68.9\%} & \textbf{25.6\%} & 5.6\% & 0.117 \\
\midrule
Recovery & LiDAR history & 58.9\% & 41.1\% & 0.0\% & 0.162 \\
& LiDAR history + recovery & 63.3\% & 33.3\% & 3.3\% & 0.163 \\
& OGM + recovery & 62.2\% & 32.2\% & 5.6\% & 0.148 \\
& \textbf{DUGM + recovery} & \textbf{76.7\%} & \textbf{13.3\%} & 10.0\% & 0.138 \\
\bottomrule
\end{tabular}
\end{table}

\textbf{External baseline comparison.}
DUGM + recovery achieves the highest success rate (76.7\%) and the
second-lowest collision rate (13.3\%) among the evaluated methods.
Nav2 MPPI attains a lower collision rate, but this is accompanied by a
61.1\% timeout rate and only 35.6\% success. ORCA and DRL-VO exhibit
substantially higher collision rates under the same sensing and actuation
conditions.

\textbf{Observation ablation.}
OGM-V2's improvement over OGM indicates that explicit velocity cues are useful. However, DDGM's lower success shows that deterministic motion projection alone does not ensure a better representation. Fixed inflation partially recovers this loss, suggesting that broader spatial support can help, yet DDGM+FI remains below OGM. DUGM achieves the strongest nominal performance. Together, these comparisons support combining motion projection with uncertainty-dependent spatial support, as implemented in DUGM, over the tested deterministic and fixed-inflation alternatives.

\textbf{Recovery ablation.}
Recovery improves success and reduces collision across all three evaluated
observation families. The largest gain occurs with DUGM, where success
increases from 68.9\% to 76.7\% and collision decreases from 25.6\% to
13.3\%, at the cost of a modest increase in timeout. Mean successful-episode
speed is not systematically reduced by recovery, suggesting that the
improvement is not obtained simply through global velocity suppression.

% ============================================================================
% Merged from sections/06_case_studies.tex
% ============================================================================
\subsection{Qualitative Case Studies}
\label{sec:qualitative-case-studies}

\begin{figure}[!t]
\centering
\includegraphics[width=\columnwidth]{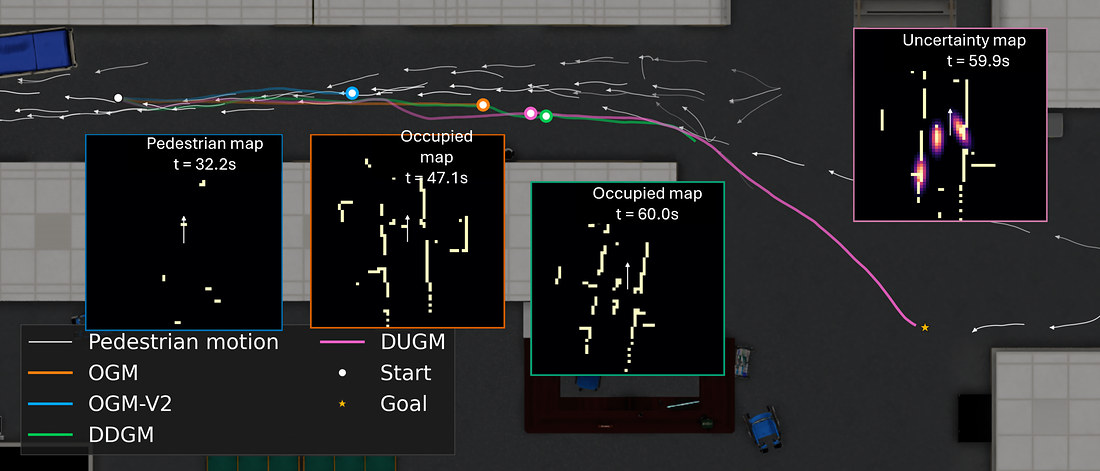}
\caption{Observation ablation: colored curves show robot trajectories for the indicated representations; white curves show pedestrian paths. Matching inset borders identify observations at locally comparable interaction states, sampled at different times.}
\label{fig:case-observation-space}
\end{figure}

Fig.~\ref{fig:case-observation-space} compares representative interactions
under OGM, OGM-V2, DDGM, and DUGM. OGM contains no explicit motion cue, while
OGM-V2 provides sparse velocity channels. DDGM projects estimated motion
deterministically, whereas DUGM expands the same rollout according to estimation
uncertainty. In the illustrated interaction, the first three policies collide,
while DUGM represents a broader future conflict region and initiates avoidance
earlier. Snapshots are selected from locally comparable interaction states rather
than identical timestamps because the policies diverge during execution.

\begin{figure}[!t]
\centering
\includegraphics[width=\columnwidth]{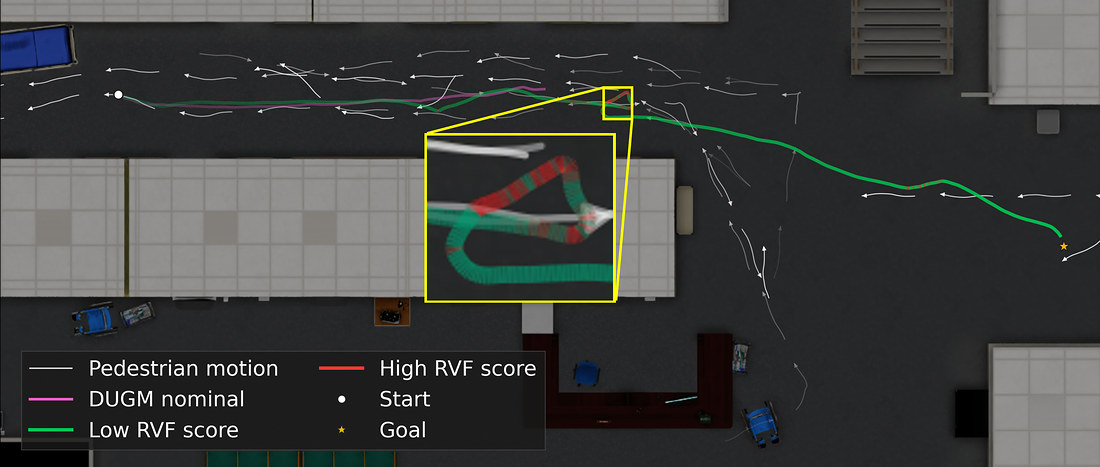}
\caption{DUGM trajectories without recovery (magenta) and with RVF-triggered recovery (colored by RVF score: green, low; red, high). White curves show pedestrian paths.}
\label{fig:case-recovery-success}
\end{figure}

Fig.~\ref{fig:case-recovery-success} illustrates a task where the nominal DUGM policy collides. The rising RVF score triggers recovery, which actively changes the robot’s local path to avoid the interaction. Once the score falls below the return threshold, nominal navigation resumes and the robot reaches the goal. This behaviour is consistent with the recovery objective: temporarily prioritizing risk reduction over immediate goal progress allows corrective manoeuvres before returning control to the nominal policy.

% ============================================================================
% Merged from sections/07_real_world.tex
% ============================================================================
\subsection{Real-Robot Evaluation}
\label{sec:experiments-real-world}

The complete controller was deployed on a TurtleBot3 Burger without policy
fine-tuning or site-specific adaptation. Observations were constructed online
from live LDS-02 LiDAR and robot odometry using the same perception interface
as in simulation; no privileged state information was available. Mean neural
inference time was $25.9\,\mathrm{ms}$ on the deployment laptop's central
processing unit (i7-13700h), indicating
that network evaluation itself fits comfortably within the nominal
$100\,\mathrm{ms}$ policy interval.

\begin{figure}[!t]
\centering
\begin{minipage}[t]{0.24\columnwidth}
  \centering\includegraphics[width=\linewidth]{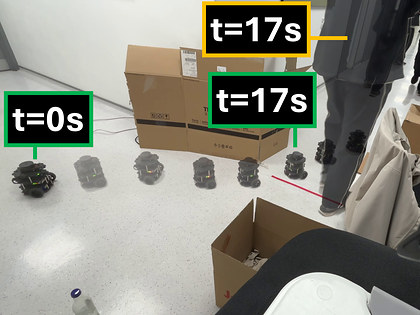}
  \panelcaption{(a) Corridor 1.}
\end{minipage}\hfill
\begin{minipage}[t]{0.24\columnwidth}
  \centering\includegraphics[width=\linewidth]{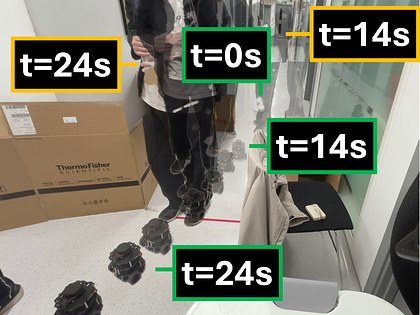}
  \panelcaption{(b) Corridor 2.}
\end{minipage}\hfill
\begin{minipage}[t]{0.24\columnwidth}
  \centering\includegraphics[width=\linewidth]{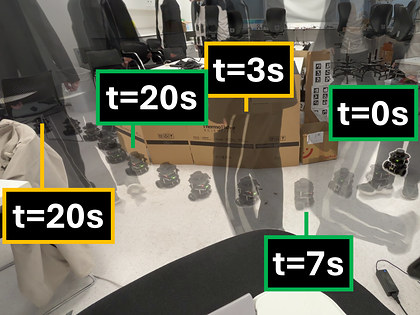}
  \panelcaption{(c) Office 1.}
\end{minipage}\hfill
\begin{minipage}[t]{0.24\columnwidth}
  \centering\includegraphics[width=\linewidth]{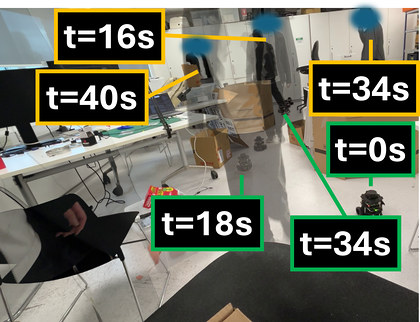}
  \panelcaption{(d) Office 2.}
\end{minipage}

\vspace{2pt}
\begin{minipage}[t]{0.24\columnwidth}
  \centering\includegraphics[width=\linewidth]{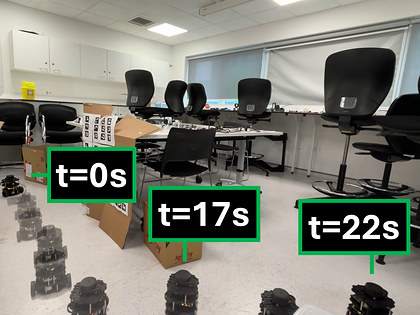}
  \panelcaption{(e) Office 3.}
\end{minipage}\hfill
\begin{minipage}[t]{0.24\columnwidth}
  \centering\includegraphics[width=\linewidth]{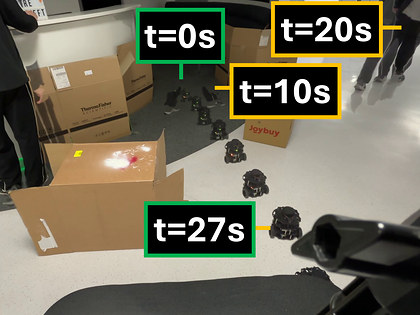}
  \panelcaption{(f) Reception 1.}
\end{minipage}\hfill
\begin{minipage}[t]{0.24\columnwidth}
  \centering\includegraphics[width=\linewidth]{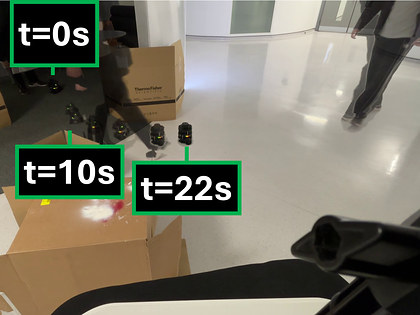}
  \panelcaption{(g) Reception 2.}
\end{minipage}\hfill
\begin{minipage}[t]{0.24\columnwidth}
  \centering\includegraphics[width=\linewidth]{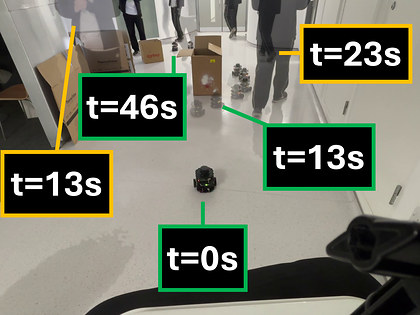}
  \panelcaption{(h) Reception 3.}
\end{minipage}
\caption{Representative views of the eight real-robot evaluation scenes with overlaid robot--pedestrian interactions at selected timestamps. Green annotations denote the robot, while yellow annotations denote pedestrians.}
\label{fig:real-world-experiments}
\end{figure}

The evaluation covered eight unseen corridor, office, and reception scenes with static and dynamic interactions Fig.~\ref{fig:real-world-experiments}.Each method was tested three times per scene with a 45 s time limit, giving 24 runs per method
Table~\ref{tab:real_robot_results}. DUGM + recovery achieved 83.3\% success and 4.2\% collision, compared with 79.2\% and 20.8\% for DUGM. Its three timeout runs subsequently reached the goal but remain classified as timeouts. These results provide preliminary evidence of direct transfer to live sensing without policy adaptation.

\begin{table}[!t]
\centering
\setlength{\tabcolsep}{4pt}
\renewcommand{\arraystretch}{1.1}
\caption{Real-robot evaluation results over 24 runs per method.}
\label{tab:real_robot_results}
\begin{tabular}{@{}lccc@{}}
\toprule
Method & SR $\uparrow$ & CR $\downarrow$ & TR $\downarrow$ \\
\midrule
DWA & 62.5\% & 8.3\% & 29.2\% \\
OGM & 70.8\% & 29.2\% & 0.0\% \\
DUGM & 79.2\% & 20.8\% & 0.0\% \\
DUGM + recovery & \textbf{83.3\%} & \textbf{4.2\%} & 12.5\% \\
\bottomrule
\end{tabular}
\end{table}

% ============================================================================
% Merged from sections/08_conclusion.tex
% ============================================================================
\section{Discussion and Limitations}
\label{sec:discussion}

The larger recovery gain with DUGM suggests a synergistic interaction between observation representation and post-training intervention. Recovery improves success by 7.8 percentage points with DUGM, compared with 4.4 points for both OGM and LiDAR history, despite DUGM's stronger nominal performance. DUGM supplies motion and uncertainty cues to nominal navigation, risk assessment, and recovery; these cues may support both risk detection and corrective action selection. The results therefore suggest that observation representation influences not only nominal performance but also the effectiveness of subsequent recovery.

The framework remains dependent on reliable online motion estimation, and recovery may query the RVF outside its nominal-rollout training distribution. Safety improvements are empirical, without formal guarantees, and real-robot evaluation remains limited in scale. Future work will strengthen motion and risk estimation and extend validation to more diverse real-world interactions.

\section{Conclusion}
\label{sec:conclusion}

This work presented a learned local-navigation framework combining DUGM with a
post-training RVF-triggered recovery policy. DUGM improved nominal navigation
over static and deterministic dynamic representations, while recovery further reduced residual
collisions across multiple observation families. The complete system achieved
76.7\% success with 13.3\% collision in the 90-episode Isaac Sim benchmark and
transferred directly to a TurtleBot3 without policy fine-tuning, achieving
83.3\% success and 4.2\% collision across 24 runs. These results support
uncertainty-aware dynamic representation and modular post-training recovery as
complementary mechanisms for robust learned local navigation.

\bibliographystyle{IEEEtran}
\bibliography{references}

\end{document}